%% file: main.tex
\documentclass[runningheads]{llncs}

\usepackage[final,year=2026,ID=10753]{eccv}
\usepackage{wrapfig}
\usepackage{eccvabbrv}
\usepackage{xcolor}
\usepackage{pgfplots}
\pgfplotsset{compat=1.18}
\usepackage{pgfplots}
\usepgfplotslibrary{groupplots}
\usepackage{graphicx}
\usepackage{booktabs}
\usepackage{subcaption}
\usepackage[accsupp]{axessibility}  % Improves PDF readability for those with disabilities.

\usepackage{siunitx}
\usepackage{booktabs}
\usepackage{pgfplots}
\pgfplotsset{compat=1.18}
\usepackage{hyperref}

\usepackage{orcidlink}

\begin{document}

\title{Limitations of Synthetic Data Generation\\in Specialized  Data-Scarce Domains} 

\titlerunning{Synthetic Data Generation in Data-Scarce Domains}

\author{
Edward Zhang\inst{1} \and
Marcel Hussing\inst{1} \and
Tanay Tandon\inst{1} \and
Shenbagaraj Kannapiran\inst{1} \and
Jason Hughes\inst{1} \and
Youkang Wang\inst{2} \and
Joshua Caswell\inst{1} \and
Agelos Kratimenos\inst{1} \and
Yi Fan Li\inst{1} \and
Milan Manoj\inst{1} \and
Ethan Sanchez\inst{1} \and
Sumukh Shrote\inst{1} \and
Camillo Jose Taylor\inst{1} \and
Daniel A. Hashimoto\inst{1} \and
Eric Eaton\inst{1}
}
\authorrunning{E. Zhang et al.}

\institute{
University of Pennsylvania, Philadelphia, PA, USA, \\
The Hong Kong Polytechnic University (PolyU), Hung Hom, Hong Kong
}
\setlength{\textfloatsep}{6pt} 
\captionsetup[table]{skip=6pt}
\raggedbottom
\maketitle

\begin{abstract}
Advances in diffusion-based generative models have motivated the use of synthetic image generation to alleviate data scarcity in vision tasks. While this strategy has shown promise in natural image benchmarks such as ImageNet, its effectiveness in sparse, high-variance real-world domains remains unclear. In this work, we focus on domains where images differ substantially from common image datasets and additional data are expensive to obtain. Against non-generative data augmentation baselines, we evaluate the downstream classifier performance improvements yielded by two schools of generative sparse data extension: distribution modeling and sample perturbation. Across five trauma classification tasks using subject-wise train--validation splits, no generative approach consistently outperforms a strong non-generative baseline. Feature-space analysis reveals recurring failure modes: memorization or collapse, distributional drift, and generation of visually plausible but simplified canonical instances that are easier to classify than real data. 
% ΤΑΝΑΥ Overall, synthetic samples reinforce existing decision boundaries rather than expand support into semantically challenging regions, suggesting that the outputs of image generation models have a pervasive underlying character that fails to meaningfully improve classifier performance in specialized domains.
Overall, in these sparse, high-variance regimes, synthetic samples tend to reinforce existing decision boundaries rather than expand support into semantically challenging regions, suggesting that current generative pipelines often fail to introduce the task-relevant variation needed to improve classifier performance beyond what strong non-generative augmentation already provides.
\end{abstract}
\input{sections/1_intro}
\input{sections/2_related_work}
\input{sections/3_background}
\input{sections/4_experiments}

\input{sections/5_analysis}
\input{sections/6_conclusion}
\input{sections/8_acknowledgements}

%\section*{Acknowledgements}
%\textit{Omitted for Blind Review}


\newpage
\bibliographystyle{splncs04}
\bibliography{main}
\end{document}

%% file: sections/1_intro.tex
\section{Introduction}
Training vision models for specialized domains, such as medical triage or industrial manufacturing, remains profoundly difficult \cite{guan2021domain, choudhary2020advancing}. These domains operate in an extreme data regime: real examples are scarce, may be restricted by ethics or privacy, may be proprietary, and are distributionally distant from the standard image datasets used to train modern vision and generative models. In such settings, standard supervised learning pipelines fail not because the task is inherently ambiguous, but because the limited available data does not capture the semantic diversity required for robust decision-making \cite{kelly2019key, shorten2019survey}.

Current state-of-the-art generative models have achieved unprecedented levels of visual fidelity in image generation  \cite{dhariwal2021diffusion}. These recent advances in image generation have motivated a growing sentiment that these models will be able to alleviate data scarcity problems by supplementing training sets with synthetic data \cite{frid2018gan}. The hypothesis is that if generative models can produce visually plausible images conditioned on class labels, then augmenting sparse real datasets with synthetic samples should \textit{improve} downstream generalization. This intuition has proven effective in natural image domains \cite{azizi2023synthetic}.

This paper tests this hypothesis in extreme data regimes, where real data is sparse, expensive to obtain, and not represented in common image datasets. In particular, we focus on trauma recognition for mass casualty incidents. Contrary to prevailing expectations, we find that diffusion-generated synthetic augmentation provides little to no improvement over conventional non-generative augmentation strategies.

Empirically, we observe three prevailing failure modes when applying generative model-based synthetic augmentation in these extreme data regimes. First, certain approaches may overfit and reproduce training examples with limited semantic variation or merely collapse into predicting a blurry average image. Second, attempts to extrapolate beyond the training distribution often yield anatomically inconsistent or semantically corrupted samples. Third, and most insidiously, even visually plausible generations frequently correspond to simplified, canonical instances of the class that are easier to classify than real data. 

In all three cases, synthetic augmentation fails to expand the effective support of the data distribution in ways that improve downstream robustness. 
% TANAY These findings suggest that the semantic variations introduced by current state-of-the-art generative models are not the variations needed to actually improve the downstream classifier performance. 
These findings suggest that, under the sparse, high-variance conditions
studied here, the semantic variations introduced by the evaluated generative
pipelines are often not the variations needed to improve downstream classifier performance.
Rather, the core challenge in extreme data regimes of ``generating good images'' goes beyond just creating more images that ``look correct'' but instead producing novel samples relevant to the task that improve the decision boundary. 

Our analysis 1) characterizes recurring failure modes of current generative
augmentation pipelines in sparse, high-variance classification settings and
2) establishes a foundation for future methods that more effectively leverage generative models to supplement these data regimes.

%% file: sections/2_related_work.tex
\section{Related Work}
\subsubsection{Synthetic Data for Improving Classification} 
Prior to the rise of generative models, synthetic data demonstrated benefits
for visual recognition systems through explicitly constructed simulation
pipelines. For example, models trained entirely on computer-rendered images can
generalize to real data when sufficient domain randomization is applied
~\cite{tremblay2018training}. By deliberately varying textures, lighting, pose,
and background during rendering, such approaches reduce overfitting to
simulator-specific artifacts and enable transfer to real imagery. Tang and
Jia~\cite{tang2023new} further study the downstream utility of synthetic data
generated using Blender, demonstrating the effectiveness of synthetic training
data when task-relevant variation can be explicitly represented and controlled
within a rendering pipeline.

More recently, learned generative models have provided an alternative means of
producing synthetic training data. Azizi et al.~\cite{azizi2023synthetic}
demonstrated that augmenting ImageNet with diffusion-generated samples yields
consistent accuracy gains across ResNet~\cite{he2016deep} and
Transformer~\cite{vaswani2017attention} classifiers. However, subsequent work
suggests that these gains are not universal. Geng et
al.~\cite{geng2024unmet} find that retrieved real images can outperform
synthetic training images for downstream classification, questioning whether
generation provides useful variation beyond that available in existing real
data. Singh et al.~\cite{singh2024synthetic} similarly examine the robustness
of classifiers trained with synthetic imagery, showing that synthetic-data
utility must be evaluated beyond performance on the nominal test distribution.

These works highlight an important distinction between synthetic data whose
variation is explicitly designed and synthetic data whose variation must be
inferred by a learned generative model. In simulation-based approaches such as
domain randomization or Blender rendering, nuisance factors and task-relevant
variation can be deliberately specified. In contrast, generative augmentation
in a sparse-data setting requires the generator to infer useful variations from
a limited empirical distribution or from its pretrained prior. Unlike ImageNet,
which provides extensive sample support, our datasets are extremely sparse and
characterized by substantial intra-class variability due to camera angle,
lighting, occlusion, background, and simulation fidelity. Thus, while synthetic
augmentation can be effective in large-scale natural-image and explicitly
parameterized simulation settings, it remains unclear when learned generative
models can supply the task-relevant variation missing from sparse,
high-variance datasets. Our work investigates this regime.

\subsubsection{Generative-Based Synthetic Data Augmentation} 
Recent advances in dif\-fu\-sion-based generative modeling have motivated their adoption for synthetic data augmentation, particularly in domains where labeled data are scarce or expensive to acquire. For example, synthetic data has been explored extensively for imaging-based subfields of medicine, such as radiology, dermatology, and ophthalmology. Guibas et al.~\cite{guibas2017synthetic} demonstrate that synthetic training data can improve performance in anatomically structured settings by expanding limited labeled datasets. More recently, Akpinar et al.~\cite{akpinar2024synthetic} investigate synthetic generation and augmentation in clinical imaging contexts, reporting performance gains when synthetic samples are incorporated into training pipelines. Sagers et al.~\cite{sagers2022improving, sagers2024augmenting} show that diffusion-based augmentation improves skin disease classification performance and enhances robustness across demographic subgroups. Similarly, Akrout et al.~\cite{akrout2023diffusion} demonstrate improved skin lesion classification with diffusion-generated samples, while Packhäuser et al.~\cite{packhauser2023generation} show that latent diffusion models can generate realistic chest radiographs suitable for downstream training.

Synthetic data has also seen widespread adoption in industrial computer vision pipelines, particularly in robotics, autonomous systems, and manufacturing, where collecting large quantities of annotated real-world data can be expensive or impractical. In these settings, simulation and generative models are frequently used to augment training data and improve robustness to environmental variability, with synthetic data and generative models being used for the purposes of anomaly detection \cite{liu2025anomaly} and real2sim transfer in domains like robotics \cite{tobin2017domain}.

However, many of these settings, while data-sparse, exhibit relatively constrained acquisition conditions. Medical, robotics, and manufacturing images are typically captured under standardized protocols, fixed camera geometries, and controlled illumination conditions \cite{zech2018variable}. In contrast, our dataset operates under extreme data scarcity combined with high nuisance variance. Samples vary substantially in viewpoint, occlusion, lighting, background clutter, and anatomical presentation, creating a substantially more fragile data manifold. Our work therefore evaluates synthetic augmentation not merely as a means of increasing sample count, but as a test of whether modern generative models can meaningfully expand semantically constrained manifolds under extreme variance.

%% file: sections/3_background.tex
\section{Data Augmentation} 

To test the efficacy of data generation methods for improving downstream image classifiers on extreme data domains, we compare a suite of non-generative and generative data generation/augmentation methods.

\subsection{Non-Generative Sparse Data Extension}

In limited-data regimes, a common strategy for improving generalization is to expand the effective training set using label-preserving transformations.  At the simplest level, image duplication can be used to mitigate class imbalance by repeating existing samples within the training set. While duplication may stabilize optimization or rebalance class frequencies, it does not introduce new visual variability and therefore provides limited benefit for improving model robustness~\cite{aghabagherloo2025impact}.

More effective approaches rely on stochastic data augmentation, which applies label-preserving transformations to increase visual diversity. We use AugMix~\cite{hendrycks2019augmix} as our primary non-generative baseline due to its strong robustness performance and computational efficiency.

\subsection{Synthetic Data Generation}
Recent advances in generative modeling have significantly expanded the frontier of synthetic data augmentation for visual recognition tasks. We summarize the two primary schools of thought regarding synthetic data generation techniques for sparse-data extension that are used in our experiments. As an additional point of comparison, we also include samples generated from a large pretrained generative model (GPT Image Generation) that does not explicitly learn anything about the seed dataset.
% TANAY
\footnote{We selected a GPT-based generator, as opposed to others such as Nano Banana 2, in part because it permits a content-moderation toggle while others do not.}

\subsubsection{Distribution Modeling}

Distribution modeling methods attempt to learn a generative approximation of the underlying data manifold and sample new images directly from this learned distribution. Our evaluation includes StyleGAN2, Stable Diffusion, and DreamBooth as representative approaches.\\[-0.5em]

\noindent{\em StyleGAN2} is a state-of-the-art GAN architecture that generates images by sampling from a learned latent distribution through a style-based generator~\cite{karras2020analyzing}.

\noindent{\em Stable Diffusion} is a text-conditioned latent diffusion model that generates images through iterative denoising in a compressed latent space~\cite{rombach2022high,zhu2023conditional}.

\noindent{\em DreamBooth} adapts a pretrained diffusion model to represent a specific subject or concept from a small set of example images by associating it with a learned identifier token~\cite{ruiz2023dreambooth} for novel image generation.

\subsubsection{Sample Perturbation}

Perturbation methods operate locally around given samples and aim to expand support in the neighborhood of real data points rather than approximating the data distribution directly.

\noindent{\em Img2Img} generates variations of an input by denoising a partially noised version of that image, preserving structure while introducing controlled variation~\cite{rombach2022high}.

\noindent{\em DreamBooth Img2Img} combines DreamBooth fine-tuning with image-to-image generation to produce variations that preserve both the input image structure and the learned concept~\cite{ruiz2023dreambooth}.

\noindent{\em IP-Adapter} conditions image generation on reference-image features rather than text, enabling image-guided generation that preserves visual characteristics while introducing variation~\cite{ye2023ip}.

\subsubsection{Large Model Sampling}

We also evaluate OpenAI's gpt-image-1.5 image generation model, which generates samples from a broad pretrained visual prior rather than a distribution learned from the seed dataset.

%% file: sections/4_experiments.tex
\section{Experimental Setup}

\subsection{Task and Dataset}

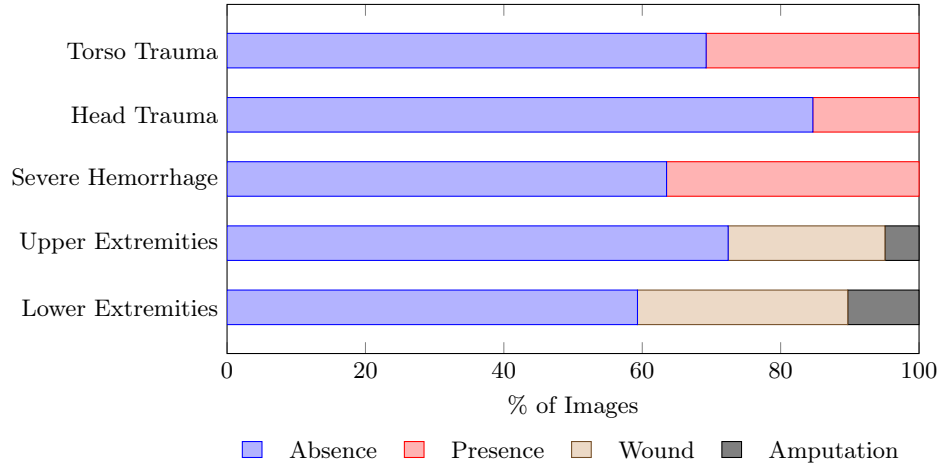
\begin{figure}[t]
\centering
\begin{tikzpicture}
\begin{axis}[
    xbar stacked,
    width=0.88\textwidth,
    height=6.2cm,
    bar width=13pt,
    xmin=0,
    xmax=100,
    xlabel={\% of Images},
    symbolic y coords={
        Lower Extremities,
        Upper Extremities,
        Severe Hemorrhage,
        Head Trauma,
        Torso Trauma
    },
    ytick=data,
    yticklabel style={
        font=\small,
        text width=3cm,
        align=right
    },
    xtick={0,20,40,60,80,100},
    xticklabel style={font=\small},
    label style={font=\small},
    enlarge y limits=0.18,
    legend style={
        at={(0.5,-0.22)},
        anchor=north,
        legend columns=4,
        font=\small,
        draw=none,
        /tikz/column sep=8pt
    },
]

\addplot coordinates {
(69.23,Torso Trauma)
(84.65,Head Trauma)
(63.51,Severe Hemorrhage)
(72.41,Upper Extremities)
(59.31,Lower Extremities)
};

\addplot coordinates {
(30.77,Torso Trauma)
(15.35,Head Trauma)
(36.49,Severe Hemorrhage)
(0,Upper Extremities)
(0,Lower Extremities)
};

\addplot coordinates {
(0,Torso Trauma)
(0,Head Trauma)
(0,Severe Hemorrhage)
(22.66,Upper Extremities)
(30.39,Lower Extremities)
};

\addplot coordinates {
(0,Torso Trauma)
(0,Head Trauma)
(0,Severe Hemorrhage)
(4.93,Upper Extremities)
(10.29,Lower Extremities)
};

\legend{Absence, Presence, Wound, Amputation}

\end{axis}
\end{tikzpicture}

\caption{Distribution of trauma labels across classification tasks. 
Torso trauma, head trauma, and severe hemorrhage are binary presence--absence
tasks, while extremity trauma additionally distinguishes wounds and amputations.}
\label{fig:trauma_distribution}
\end{figure}

We evaluate synthetic data augmentation for trauma recognition within the
PRONTO robotic triage system developed for the DARPA Triage Challenge
\cite{hughes2025multi}. The system is designed to perform remote primary
triage in mass-casualty incidents using heterogeneous aerial and ground robotic platforms. In this work, we focus specifically on the visual trauma recognition component, considering severe hemorrhage, head trauma, torso trauma, upper extremity trauma, and lower extremity trauma.

The dataset consists of images captured during realistic high-fidelity disaster simulations and institutional data collection efforts using actors and medical manikins in moulage (makeup simulating injuries).  All human subjects data was collected and used under IRB review and approval (UPenn IRB \#856195 and USAISR Protocol Number \#M-11034). The images depict simulated traumatic injuries across a wide range of environmental conditions, including variation in background, lighting, time of day, and camera viewpoint. The initial set of 566 images was manually filtered through multiple human review passes to remove ambiguous labels and low-quality images, producing the final curated dataset of 362 images of 211 unique subjects, with some subjects photographed from up to four distinct camera angles. To avoid subject-level leakage between training and evaluation sets, all train--validation splits are performed at the subject level rather than the image level. For each of the five trauma classification tasks, we construct one fixed 70/30 subject-wise train--validation split independently, since label distributions differ across trauma categories. For context on acquisition cost, real trauma data collection with physical manikins and moulage typically yielded only 8--16 usable images per roughly four hours of setup and capture, whereas every locally evaluated generation pipeline produced at least 2{,}000 synthetic images per class in under 48 GPU-hours on a single NVIDIA B200 or A6000.

Prior to classification, images are preprocessed to reduce background variability and focus the model on the human subject. We apply Grounding DINO \cite{liu2024grounding} to detect and crop tighter human-centered bounding boxes around each subject. This preprocessing step removes large portions of irrelevant background context while preserving the visual features associated with injury appearance.

\subsection{Classification Pipeline}

For each trauma type, images are encoded using a frozen DINOv2 backbone~\cite{oquab2023dinov2}. The resulting feature embeddings are used to train a three-layer multilayer perceptron (MLP) classifier. 
% TANAY Performance is evaluated using unweighted precision on the held-out validation set of real images. This metric is chosen because the dataset exhibits significant class imbalance (Fig.~\ref{fig:trauma_distribution}), making accuracy an unreliable performance measure.
Performance is evaluated using macro-recall on the held-out validation set of real images. The dataset exhibits significant class imbalance (Fig.~\ref{fig:trauma_distribution}), which makes accuracy an unreliable measure. While precision, F1-score, and per-class analysis are standard in imbalanced settings, our validation set is both small and imbalanced, so these metrics become highly unstable in this regime. We therefore report macro-recall, which remains comparatively stable and interpretable.

The independent variable across experiments is the augmentation strategy used to expand the training set. In all cases, the original training split serves as the seed dataset, which is expanded until each class contains 2,000 training samples. 
% TANAY For each augmentation method, this expansion process is repeated three times with different random seeds to estimate variance across runs.
For stochastic augmentation methods, training-set expansion is repeated across three independent random seeds while evaluation is performed on the same fixed validation split for each trauma task. For generative approaches, these runs correspond to independently generated synthetic datasets, while AugMix is repeated with independent augmentation seeds. Image duplication is deterministic for a fixed training split and therefore exhibits zero variance across repetitions. Consequently, the reported standard errors reflect augmentation or generation stochasticity rather than variation in validation-set composition.

\subsection{Data Augmentation Experiments}

To evaluate the impact of synthetic data augmentation, we compare three classes of dataset expansion strategies: non-generative baselines, distribution modeling methods, and sample perturbation methods. In all experiments, the original training split serves as the seed dataset and is expanded until each class contains 2000 training samples.

\paragraph{Non-Generative Baselines.}
We first evaluate standard label-preserving augmentation strategies that do not involve generative modeling. Here, Image duplication and AugMix~\cite{hendrycks2019augmix} provide a strong, essentially computationally ``free'', non-generative augmentation baseline.

\paragraph{Distribution Modeling Methods.}
For distribution modeling approaches, each generative model is trained or adapted using the seed training split and then used to generate synthetic samples until each class contains 2000 images. StyleGAN is trained directly on the seed images and samples are generated by drawing latent vectors from the learned distribution. For Stable Diffusion, we finetune each checkpoint such that its caption should produce its associated training image. For DreamBooth, the class label in the prompt is replaced with a unique identifier token that is fine-tuned to represent the visual concept present in the seed images. Each method, once trained, is sampled 2000 times per label to get 2000 samples per class label. ChatGPT API Image Generation was queried with 2000 computationally generated unique prompts to generate images whose subjects varied by age, gender, and background.

\paragraph{Sample Perturbation Methods.}
For sample perturbation approaches, each real training image is used as a seed to generate multiple variants until the total number of samples per class reaches 2000. Img2Img editing generates variants by prompting each training image with its own caption multiple times. DreamBooth + img2img does the same thing, but this time with the caption's embedded label being replaced with the learned unique token. IP-Adapter generates variants by prompting each seed image with the embedding of itself multiple times.

\paragraph{On Tuning and Comparability.}
Our goal is not exhaustive per-method optimization but evaluation under standardized, practitioner-aligned settings that reflect realistic usage. Because these approaches rely on fundamentally different priors and operating mechanisms, there is no unified notion of optimal tuning that would enable a strictly controlled comparison. This also accounts for the differing treatment of nuisance variation: the GPT generator cannot be fine-tuned, so such variation must be introduced explicitly through prompts, whereas all other baselines learn it implicitly through fine-tuning on our own dataset.

%% file: sections/5_analysis.tex
\section{Results and Analysis}
\subsection{Qualitative Analysis}
Img2Img diffusion editing produced variations that largely preserved the structure of the original images while introducing small local changes such as minor shifts in texture, lighting, or background elements. As a result, the generated samples often appeared visually coherent but exhibited relatively limited diversity compared to other methods. In contrast, IP-Adapter generation produced substantially greater diversity in camera angle, background, and subject pose, though this occasionally came at the cost of visual coherence, with some generated samples containing anatomical inconsistencies or semantic irrelevance. Img2Img with DreamBooth components produced results that fell between these two behaviors.

For distribution modeling approaches, several distinct failure modes were observed across the evaluated methods. StyleGAN frequently exhibited signs of mode collapse or dataset memorization, producing images that closely resembled specific training samples and suggesting limited coverage of the underlying data distribution. Stable Diffusion generations demonstrated greater visual diversity, but often failed to maintain semantic consistency with the trauma scenarios represented in the dataset. DreamBooth-based generation produced the most visually coherent samples among the distribution modeling approaches; however, these images tended to be highly similar to one another, indicating only limited expansion of the underlying data manifold. GPT-generated images, while appearing realistic and high-fidelity at first glance, displayed a comparable level of visual homogeneity to DreamBooth outputs, suggesting that they likewise provide only modest increases in effective data diversity.

\begin{figure}[t!]
    \centering
    \includegraphics[width=.9\linewidth]{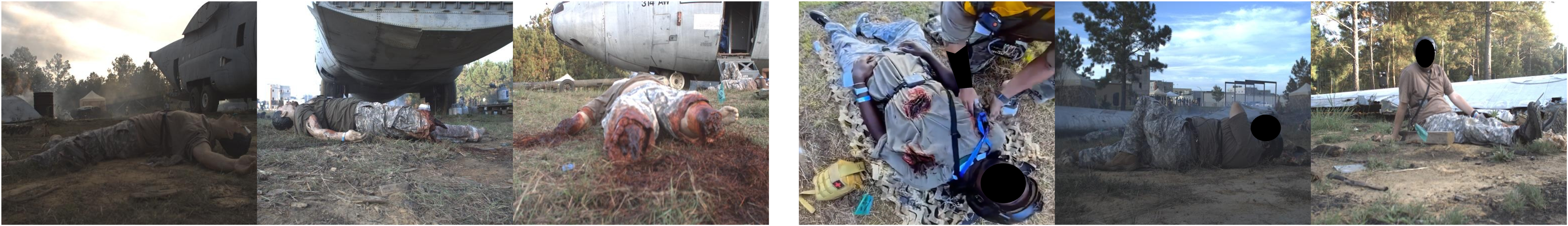}
    \caption{Example images from realistic medical trauma simulations, including medical manikins (left  images) and de-identified actors in moulage (right images) - (UPenn IRB \#856195, USAISR Protocol \#M-11034). }
    \label{fig:real_data}
    \vspace{-1em}
\end{figure}
\begin{figure}[t!]
    \includegraphics[width=\linewidth]{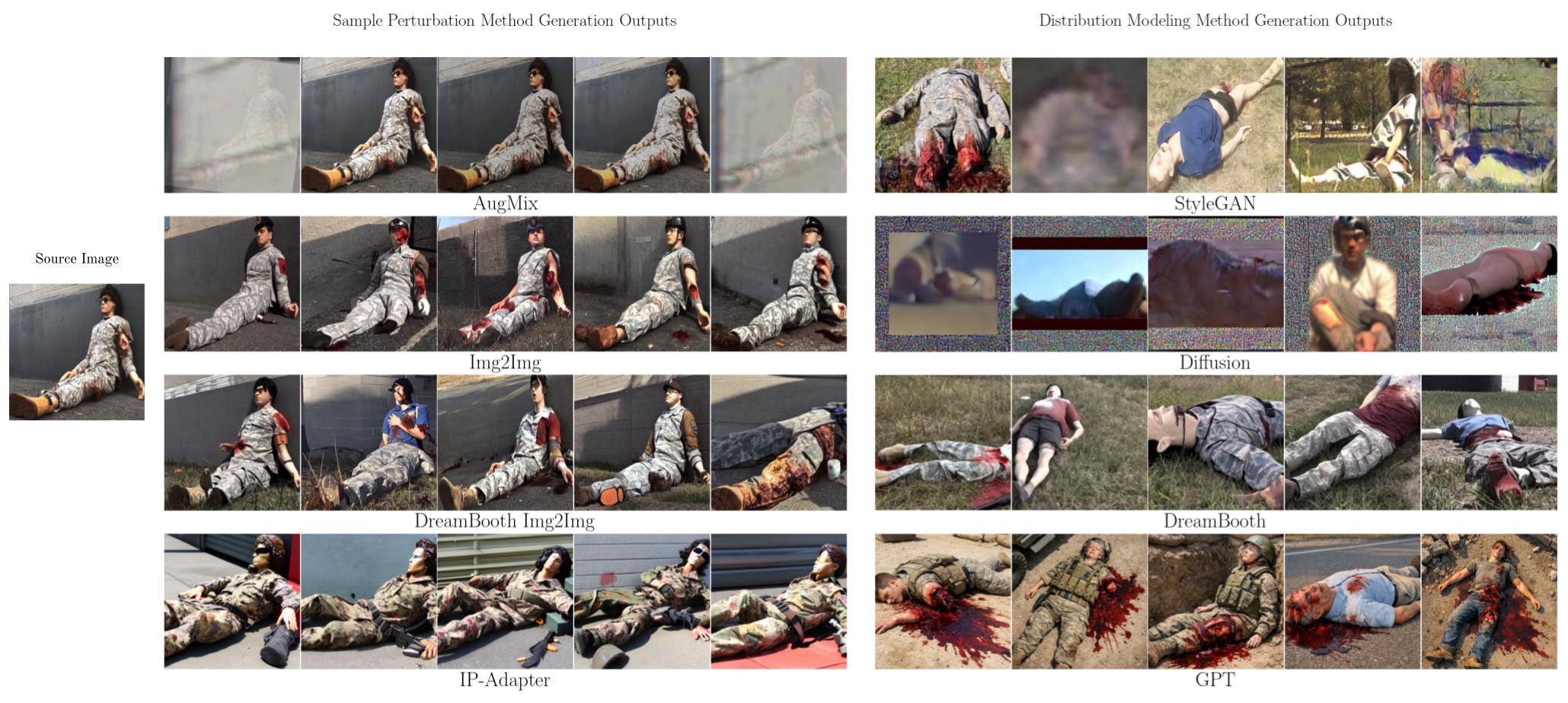}
    \caption{Generative Model Outputs. Compared to the real images in Figure~\ref{fig:real_data}, many of the distribution model images are unrealistically clean regarding lack of adverse lighting, dirt and/or grime obscuring the wound, and potentially suboptimal camera angles, especially from GPT.}
    \label{fig:generation_gallery}
\end{figure}

\subsection{Quantitative Analysis}
\begin{table}[t]
\centering
\caption{Macro-recall (\%) with standard error.}
\label{tab:recall_with_std}
\begin{tabular}{lccccc}
\toprule
Method & Sev.~Hem. & Head & Torso & Upper Ext. & Lower Ext. \\
\midrule
Real Data Only & 50.00{\scriptsize$\pm$0.00} & 50.00{\scriptsize$\pm$0.00} & 50.00{\scriptsize$\pm$0.00} & 33.33{\scriptsize$\pm$0.00} & 33.33{\scriptsize$\pm$0.00} \\
Image Dup & 63.11{\scriptsize$\pm$0.00} & 44.15{\scriptsize$\pm$0.00} & \textbf{67.73}{\scriptsize$\pm$0.00} & 39.37{\scriptsize$\pm$0.00} & 60.81{\scriptsize$\pm$0.00} \\
AugMix (Baseline)               & 63.38{\scriptsize$\pm$1.62} & 48.61{\scriptsize$\pm$0.45} & 62.45{\scriptsize$\pm$0.30} & 37.78{\scriptsize$\pm$0.76} & \textbf{66.56}{\scriptsize$\pm$1.26} \\
\midrule
IP-Adapter           & 56.00{\scriptsize$\pm$2.14} & 49.37{\scriptsize$\pm$0.79} & 55.36{\scriptsize$\pm$0.97} & 42.87{\scriptsize$\pm$0.67} & 46.13{\scriptsize$\pm$2.44} \\
Img2Img               & 49.75{\scriptsize$\pm$0.59} & 54.04{\scriptsize$\pm$0.74} & 60.55{\scriptsize$\pm$1.01} & 40.32{\scriptsize$\pm$1.09} & 47.16{\scriptsize$\pm$0.49} \\
DB Img2Img   & 57.76{\scriptsize$\pm$0.71} & 51.36{\scriptsize$\pm$1.24} & 63.69{\scriptsize$\pm$0.19} & 41.11{\scriptsize$\pm$0.67} & 57.06{\scriptsize$\pm$1.51} \\
\midrule
StyleGAN              & 54.13{\scriptsize$\pm$1.18} & 56.27{\scriptsize$\pm$0.02} & 58.41{\scriptsize$\pm$0.12} & 27.43{\scriptsize$\pm$1.33} & 17.49{\scriptsize$\pm$0.00} \\
Diffusion     & 48.45{\scriptsize$\pm$1.24} & 54.34{\scriptsize$\pm$1.10} & 61.33{\scriptsize$\pm$0.37} & \textbf{44.12}{\scriptsize$\pm$0.95} & 31.20{\scriptsize$\pm$0.52} \\
DreamBooth            & 58.46{\scriptsize$\pm$2.67} & 52.48{\scriptsize$\pm$0.54} & 60.04{\scriptsize$\pm$0.86} & 42.08{\scriptsize$\pm$0.71} & 49.66{\scriptsize$\pm$1.11} \\
\midrule
GPT                   & \textbf{82.50}{\scriptsize$\pm$0.00} & \textbf{56.91}{\scriptsize$\pm$0.33} & 47.00{\scriptsize$\pm$0.70} & 40.57{\scriptsize$\pm$1.39} & 40.72{\scriptsize$\pm$0.54} \\
\bottomrule
\end{tabular}
\end{table}

Table~\ref{tab:recall_with_std} reports classifier performance for each augmentation strategy across all trauma categories. When trained on the sparse real dataset without augmentation, the classifiers collapse to predicting the majority class, yielding macro-recall of 50\% for
the binary hemorrhage, head, and torso tasks and 33.33\% for the three-class
extremity tasks. 
While several synthetic augmentation methods improve performance over the real-data-only baseline, none consistently outperform the AugMix baseline across all categories. In particular, generative approaches exhibit highly variable behavior: certain methods achieve improvements in specific categories, but these gains are not consistent across tasks. It is also important to note that StyleGAN's performance is somewhat misleading, as its dataset memorization may inflate recall by effectively behaving like image duplication rather than by producing useful novel samples.
% TANAY
A pixel-space nearest-neighbor analysis supports this interpretation: in the hemorrhage-presence class, 19.9\%, 13.3\%, and 3.1\% of generated samples fall below the real-data p10, p5, and p1 nearest-neighbor MSE thresholds, respectively, and 1.28\% lie closer to a training image than any pair of real training images lies to each other---indicating a tail of near-duplicate samples consistent with localized memorization.
On the other hand, the unusually high performance for GPT on the hemorrhage category likely reflects the generation of visually exaggerated canonical injury patterns that are easier for the classifier to detect.
This inconsistency suggests that the effectiveness of synthetic augmentation may depend on how closely generated samples align with the underlying data distribution. To better understand these differences, we analyze the relationship between synthetic and real data distributions in feature space. Specifically, we examine the distance between generated and real image embeddings to determine whether distributional alignment correlates with downstream classifier performance.

\begin{figure}

    \centering
    \includegraphics[width=\linewidth]{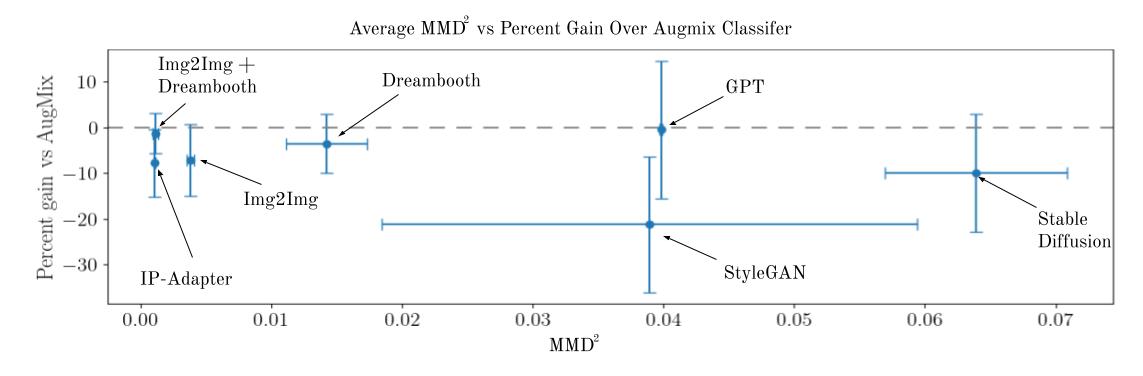}
    \caption{MMD² between each synthetic dataset and the real data versus the average percent gain in classifier recall over the AugMix baseline. Whiskers denote standard error. Sample-perturbation methods tend to remain closer to the real data distribution and yield larger gains, while distribution-modeling approaches lie farther away. Overall, greater distributional similarity correlates with larger performance improvements}
    \label{fig:mmd}
\end{figure}

\vspace{-15pt}
\subsubsection{Feature Space Analysis}
In order to quantify the relationship between synthetic data alignment and classifier performance, we compute the Maximum Mean Discrepancy (MMD$^2$) between DINOv2 feature embeddings of generated samples and real images. Figure~\ref{fig:mmd} plots this feature-space distance against the percentage recall gain relative to the AugMix baseline. As expected, lower MMD$^2$ values generally correspond to higher performance gains. Perturbation-based methods consistently produce synthetic samples that remain closer to the real data distribution, whereas distribution modeling approaches such as StyleGAN and Stable Diffusion generate samples that lie further from the empirical feature manifold and therefore tend to yield lower gains. DreamBooth generation lies between these regimes: its samples remain closer to the real distribution than those of StyleGAN or Stable Diffusion, but not as consistently aligned as those produced by perturbation-based methods.

\paragraph{t-SNE Observations.}
To further visualize these relationships, we generate t-SNE projections of the feature embeddings with synthetic samples overlaid on the real data distribution. These projections of the DINOv2 feature embeddings reveal clear differences in how each augmentation strategy populates the feature space relative to the real data distribution. AugMix samples form tight clusters around existing real data points, consistent with its design as a transformation-based augmentation method that produces local variations of the original images.
\begin{figure}
    \centering
    \includegraphics[width=\linewidth]{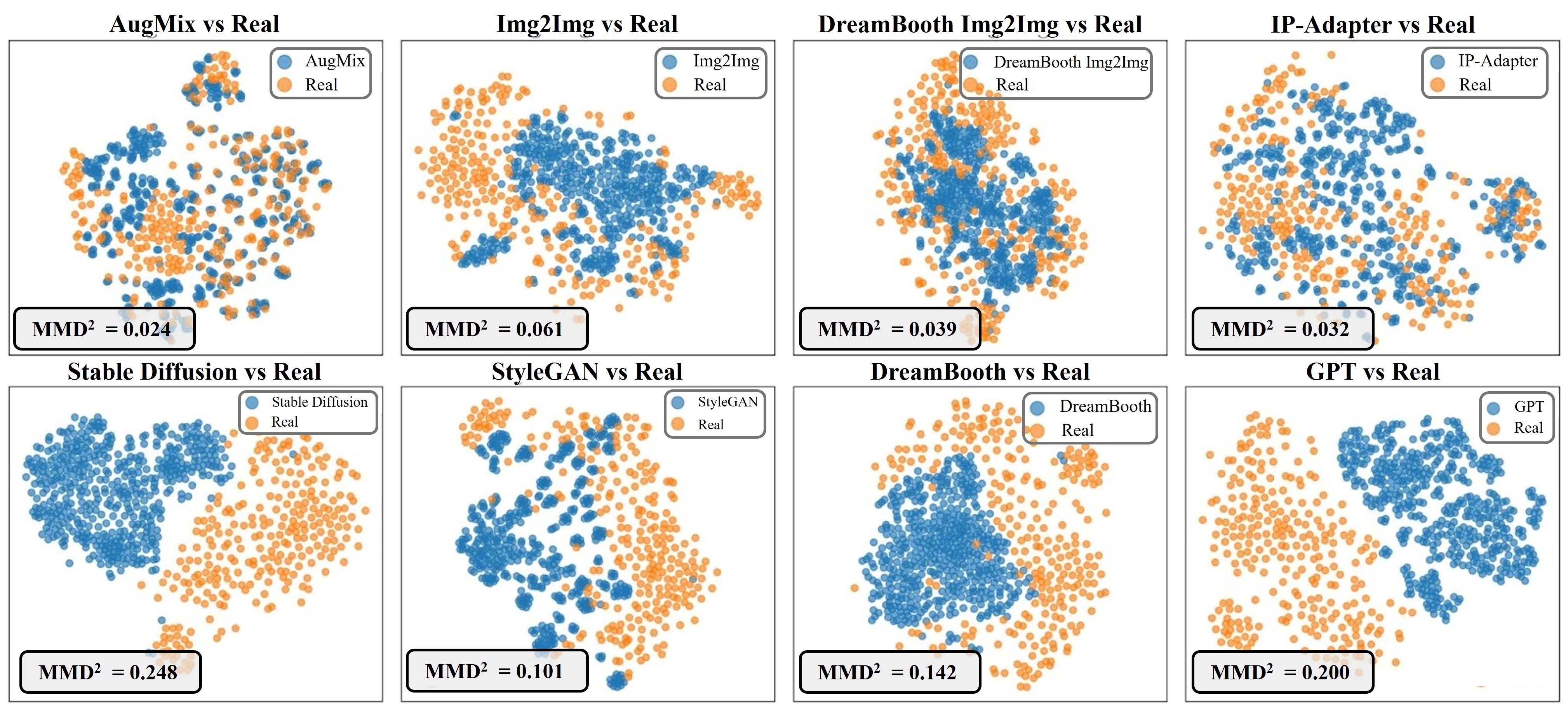}
    \caption{t-SNE plots of Generated Images Relative to Real Data - Sample Perturbation Methods tend to have lower $MMD^2$ and fill in gaps between existing real data while the Distribution Modeling Methods tend to spawn their own separate cluster, resulting in higher $MMD^2$}
    \label{fig:tsne}
\end{figure}
StyleGAN exhibits two distinct behaviors: a subset of generated samples lies directly on top of existing real images, suggesting memorization of training examples, while another group forms a separate cluster corresponding to visually blurred or averaged outputs that are largely disconnected from the real data manifold. Stable Diffusion shows an even stronger distributional shift, with most generated samples occupying regions of feature space that have little overlap with the real data.

In contrast, perturbation-based methods tend to populate the intermediate regions between existing real samples. Rather than forming separate clusters, these methods appear to expand the local support of the empirical data manifold by filling gaps between real observations. Direct DreamBooth generation produces samples that remain largely contained within the overall envelope of the real distribution; however, these samples cluster tightly together, indicating limited diversity and a tendency to generate visually similar outputs.

\subsection{Follow-Up Experiments}
Motivated by the observations from the primary experiments and feature-space analysis, we conducted additional studies to better understand the role synthetic data plays in classifier performance.

\begin{table*}[t]
\centering
\scriptsize
\begin{tabular}{lccccc}
\toprule
\textbf{Method} & \textbf{Hemorrhage} & \textbf{Head} & \textbf{Torso} & \textbf{Upper Ext.} & \textbf{Lower Ext.} \\
\midrule

Diffusion 
& \textcolor{red}{-4.6}
& \textcolor{green!50!black}{+0.1}
& \textcolor{red}{-4.6}
& \textcolor{red}{-5.1}
& \textcolor{red}{-9.5} \\

DreamBooth 
& \textcolor{red}{-8.2}
& \textcolor{green!50!black}{+0.1}
& \textcolor{red}{-2.8}
& \textcolor{red}{-2.6}
& \textcolor{green!50!black}{+2.3} \\

GPT 
& \textcolor{red}{-15.2}
& \textcolor{red}{-7.9}
& \textcolor{red}{-23.1}
& \textcolor{red}{-26.9}
& \textcolor{green!50!black}{+4.7} \\

IP-Adapter 
& \textcolor{green!50!black}{+2.1}
& \textcolor{green!50!black}{+32.6}
& \textcolor{green!50!black}{+6.8}
& \textcolor{red}{-3.6}
& \textcolor{red}{-4.2} \\

SD Img2Img 
& \textcolor{red}{-9.0}
& \textcolor{green!50!black}{+8.0}
& \textcolor{green!50!black}{+7.8}
& \textcolor{red}{-9.7}
& \textcolor{green!50!black}{+21.8} \\

SD Img2Img + DB 
& \textcolor{green!50!black}{+1.2}
& \textcolor{green!50!black}{+9.3}
& \textcolor{green!50!black}{+8.9}
& \textcolor{red}{-0.9}
& \textcolor{green!50!black}{+2.7} \\

StyleGAN 
& \textcolor{red}{-7.0}
& \textcolor{green!50!black}{+2.5}
& \textcolor{green!50!black}{+8.6}
& \textcolor{red}{-20.5}
& \textcolor{green!50!black}{+5.3} \\

\bottomrule
\end{tabular}

\caption{Absolute percentage-point change in macro-recall relative to the AugMix baseline for end-to-end ResNet classifiers trained with different synthetic augmentation pipelines.}
\label{tab:delta_augmix_resnet}
\end{table*}

\subsubsection{Sensitivity to Classifier Architecture and Mixing Ratio}
Our primary experiments use a frozen DINOv2 backbone with an MLP head, so a
natural concern is whether the observed behavior is specific to that pipeline.
To test this, we repeat the core comparison under end-to-end ResNet50 training.
As shown in Table~\ref{tab:delta_augmix_resnet}, no synthetic pipeline
consistently outperforms the AugMix baseline across categories, mirroring the
frozen-backbone results.

\begin{table}[b]
\centering
\begin{tabular}{lcccc}
\toprule
Method & 1:1 & 1:2 & 1:5 & 1:10 \\
\midrule
GPT         & 56.6 & \textbf{59.3} & 55.5 & 51.2 \\
IP-Adapter  & \textbf{62.1} & 61.8 & 61.2 & 58.0 \\
SD Img2Img  & 59.6 & 57.6 & \textbf{60.6} & 58.2 \\
\bottomrule
\end{tabular}
\caption{Mean macro-recall (\%) across the five trauma classification
tasks under different real-to-synthetic training ratios. Bold indicates
the best ratio for each generation method.}
\label{tab:ratio}
\end{table}

We additionally evaluate sensitivity to the amount of synthetic data introduced during training. For GPT, IP-Adapter, and Stable Diffusion Img2Img, we vary the real-to-synthetic sampling ratio across 1:1, 1:2, 1:5, and 1:10. Table~\ref{tab:ratio} reports macro-recall averaged across the five trauma classification tasks. No common optimal ratio emerges across generation methods: GPT performs best at 1:2, IP-Adapter at 1:1, and Img2Img at 1:5. Performance also varies substantially across individual trauma categories, indicating that simply increasing the relative quantity of synthetic data does not reliably improve downstream performance. Instead, the utility of synthetic augmentation remains sensitive to both the generation method and the downstream task.

\subsubsection{Testing Hardness of Synthetic Samples}

A possible explanation for the limited gains from synthetic augmentation is that generated images may not meaningfully expand the classifier’s decision boundary. We investigate this by evaluating whether synthetic samples represent challenging examples relative to real data. For each synthetic generation method (excluding StyleGAN and Stable Diffusion due to memorization and generation failures), we manually hand-select approximately 50 curated exemplar images and evaluate them using classifiers trained on AugMix-augmented real data. If synthetic samples meaningfully expanded the data manifold, we would expect these classifiers to struggle on such images. Instead, classifiers all achieved higher recall on synthetic exemplars than on matched real-data exemplars. Depending on the generation method, recall was between 11\% and 45\% higher on synthetic images, with DreamBooth and Img2Img producing the largest gains. 

\begin{figure}
    \centering    \includegraphics[width=.85\linewidth]{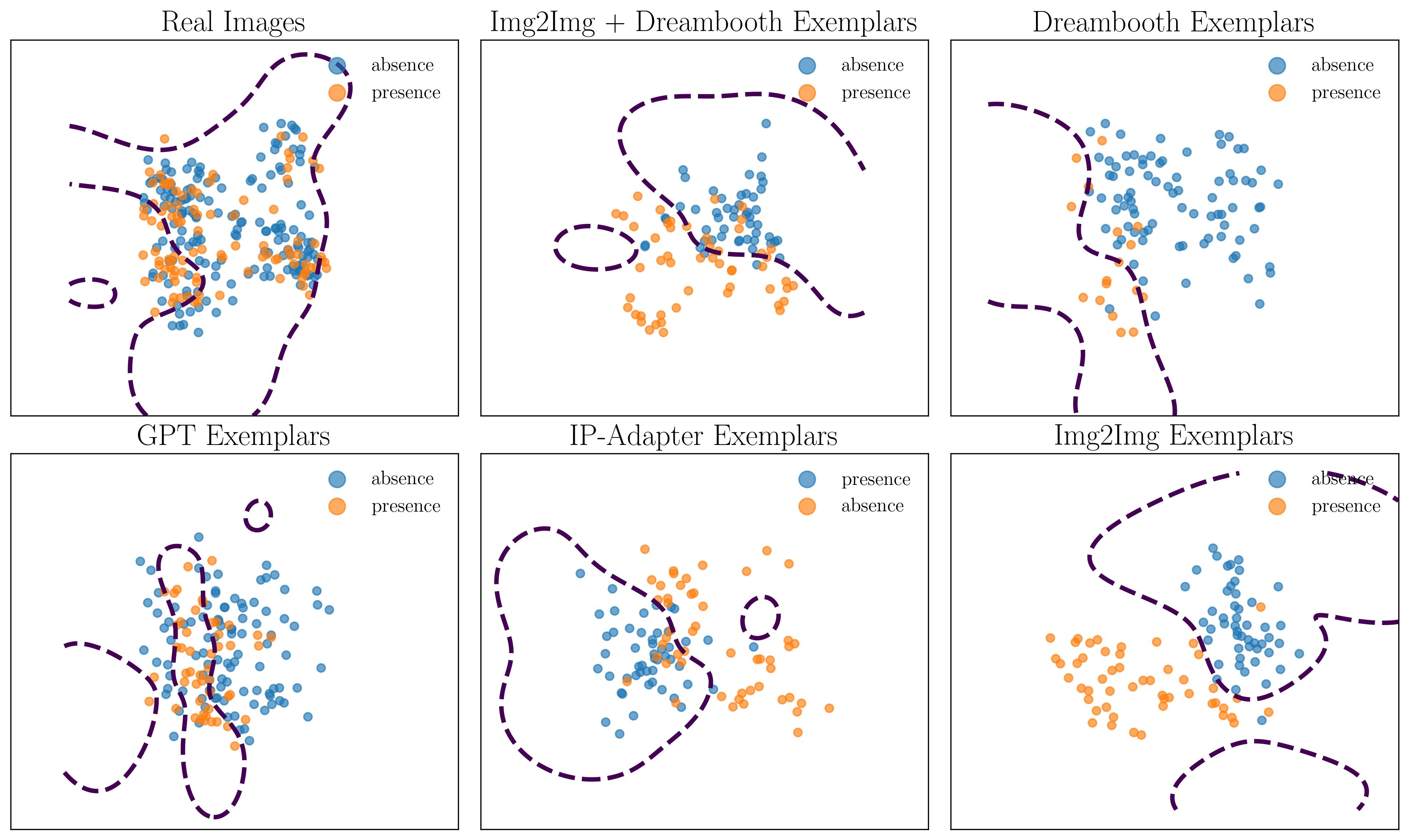}
    \caption{DINOv2 embeddings, projected onto the top two principal components of the real training data, show that synthetic exemplars are more easily separable than real images, requiring simpler classification boundaries, as illustrated by a SVM with an RBF kernel} 
\label{fig:exemplar_dino}

\end{figure}

This suggests that generated images tend to represent simplified or canonical instances of each class rather than challenging decision-boundary cases. Rather than introducing novel or difficult examples, these methods primarily generate samples that lie within already well-modeled regions of the feature space. This behavior becomes particularly clear in Figure ~\ref{fig:exemplar_dino}{} when visualizing the exemplar samples in feature space. 
 % I think its arguable that the GPT boundary is still estronger than the Real boundary. it's not as pronounced, but there is a very clear orange neighborhood inside

\subsubsection{Evaluation in a Less Sparse Regime}
One possible explanation for the limited effectiveness of synthetic augmentation in our primary experiments is the extreme sparsity of the trauma dataset. To test whether this behavior persists in a less sparse setting, we repeat the experimental protocol on a larger dataset drawn from a different domain: the Intel Robotic Welding Dataset.

\begin{figure}
    \centering
    \includegraphics[width=0.6\linewidth]{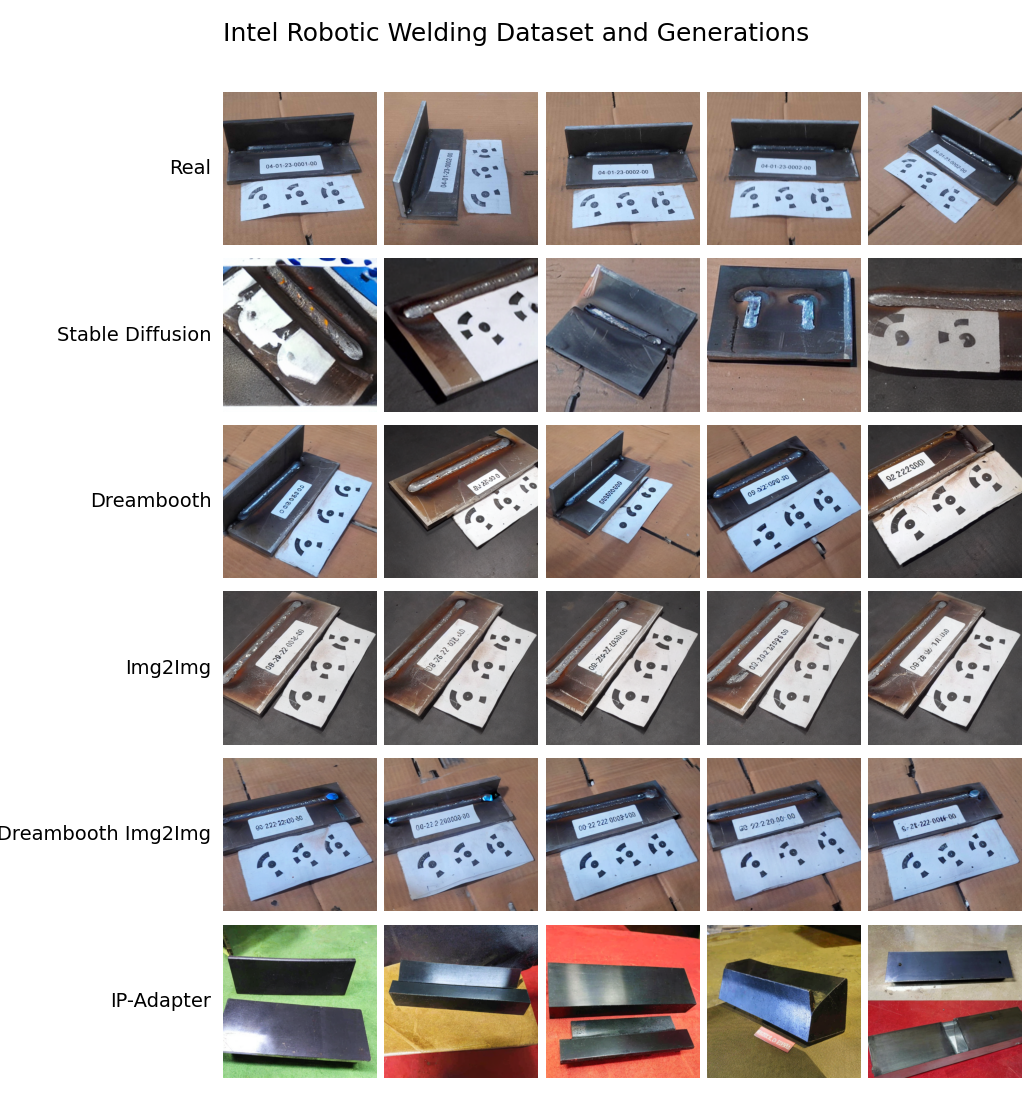}
    \caption{Examples of the Synthetic Generation Outputs for the Welding Data}
    \label{fig:weld_gallery}
\end{figure}

The Intel Welding Dataset \cite{stemmer2024unsupervised} contains over 4,000 weld instances spanning 12 quality categories (one ``Good'' class and multiple defect types) across $\sim$20,000 images. For our experiments, we collapse the individual defect categories into a single ``Defect'' class and compare them against the ``Good'' class, yielding a binary weld-quality classification task. Examples of outputs generated by our different methods can be seen in \ref{fig:weld_gallery}. Compared to the trauma dataset, this dataset provides substantially greater sample support while still representing a specialized visual domain unlikely to be well represented in the pretraining data of diffusion models.

We first train our DINO-based classifier using 2,000 real images per class to establish a baseline. We then replicate the augmentation procedure by using 500 real images per class as seed data for AugMix and each generative method (DreamBooth, Stable Diffusion, Img2Img, IP-Adapter, and DreamBooth+Img2Img), generating synthetic samples until each class again reaches 2,000 images.

Despite the increased dataset size, the results mirror those observed in the trauma experiments: synthetic generation methods do not consistently outperform AugMix. This suggests that the limitations observed earlier are not solely a consequence of extreme sparsity, but may instead reflect broader challenges in using generative models to meaningfully expand the effective training distribution.

\begin{table}[t]
\centering
\begin{tabular}{lc}
\toprule
\textbf{Augmentation Method} & \textbf{Macro-recall (\%)} \\
\midrule
Real Data Only & 77.48 \\
AugMix (baseline) & \textbf{80.46} \\
\midrule
DreamBooth (Direct Gen) & 77.20 \\
Stable Diffusion (FT) & 75.41 \\
\midrule
Img2Img & 77.79 \\
Img2Img + DreamBooth & 76.62 \\
IP-Adapter & 75.67 \\
\bottomrule
\end{tabular}
\caption{Macro-recall on weld classification with synthetic augmentation.}
\label{tab:weld_results}
\end{table}

\begin{table}[b]
\centering
\begin{tabular}{lrrrrr}

Data Generation Method & Hemo & Head & Torso & Upper Ext & Lower Ext \\
\hline
DreamBooth & {\color{red}-7.02\%} & {\color{red}-2.25\%} & {\color{red}-9.44\%} & {+17.57\%} & {\color{red}-15.94\%} \\
Img2Img & {+5.31\%} & {+3.94\%} & {+1.12\%} & {+23.89\%} & {\color{red}-9.37\%} \\
Img2Img + DreamBooth & {+1.63\%} & {+9.06\%} & {\color{red}-4.41\%} & {\color{red}-1.61\%} & {\color{red}-4.41\%} \\
IP-Adapter & {+4.93\%} & {+4.37\%} & {\color{red}-4.33\%} & {+17.69\%} & {\color{red}-4.22\%} \\
GPT & {\color{red}-6.25\%} & {+4.65\%} & {\color{red}-7.36\%} & {\color{red}-14.17\%} & {\color{red}-5.41\%} \\
\hline
\end{tabular}
\caption{Performance gain relative to AugMix when exemplar synthetic data are mixed with real training data across injury categories. Positive values indicate improvement over AugMix; negative values (red) show that adding synthetic data hurts performance.}
\label{tab:exemplar_augmix_gain}
\end{table}

\subsubsection{Mixing Curated Synthetic Exemplars with Real Data}
One possible explanation for the limited gains from fully synthetic augmentation is that automatically generated samples may include low-quality or semantically inconsistent images. To test whether high-quality synthetic exemplars could provide targeted benefit, we used the manually curated subset of approximately 50 visually coherent and label-consistent exemplars from each generation method.

These curated exemplars were first incorporated directly into the real training set. However, classifiers trained on this mixed dataset tended to collapse toward predicting the majority class. To mitigate this effect, we instead applied AugMix on top of the combined set of real images and curated exemplars, allowing the augmented training pipeline to operate over both real and synthetic samples.

The results indicate that carefully curated synthetic samples can modestly extend the effective support of the data distribution beyond what strong non-generative augmentation alone provides, especially when mixed with real data. That said, the method of focusing these generative methods to only generate exemplar level images remains an open question.

%% file: sections/6_conclusion.tex
\vspace{-15pt}

\section{Conclusions}
\vspace{-10pt}
By investigating whether modern generative models can meaningfully improve classifier performance in extreme low-data regimes, we find that, across a diverse set of distribution modeling and perturbation-based synthetic data generation strategies, synthetic augmentation does not consistently outperform strong non-generative baselines such as AugMix.

Feature-space analysis reveals that the effectiveness of synthetic augmentation is tied to how well generated samples align with the real data distribution, but further analysis suggests that when synthetic images \textit{are} aligned feature-wise with the real data \textit{and} semantically correct, they tend to be inherently easier classification cases than real-world samples, indicating that these generated samples typically occupy already well-modeled regions of feature space rather than introducing novel or challenging examples that expand the classifier’s decision boundary. We also show that this is not limited to medical triage, as replicating the experiments on the larger Intel Robotic Welding dataset yields similar results, suggesting that the limitations of synthetic augmentation are not solely a consequence of extreme sparsity. Experiments using large-scale proprietary image generation models likewise fail to produce substantial gains over strong augmentation baselines. Even when manually curated synthetic exemplars are incorporated into the training pipeline, improvements remain modest and are primarily observed for perturbation-based generation methods. 

% TANAY Taken together, these results suggest that while modern generative models can produce visually convincing images, they do not yet reliably generate samples that expand the effective training distribution in ways that meaningfully improve classifier performance in real-world deployment scenarios.
Taken together, these results suggest that while modern generative models can produce visually convincing images, they do not yet reliably generate samples that expand the effective training distribution in ways that meaningfully improve classifier performance in these specialized, data-scarce regimes.
Instead, simple transformation-based augmentation methods remain surprisingly competitive—and in many cases preferable—in sparse, high-variance domains.

\subsection{Limitations}

This study has several limitations that constrain the generality of our
conclusions. First, our experiments focus primarily on trauma recognition,
with the Intel Robotic Welding Dataset serving as a second specialized domain.
Although these datasets differ substantially in acquisition process, scale, and
visual content, both are appearance-driven image classification tasks. The
welding experiments therefore provide supporting evidence that the observed
behavior is not unique to the trauma dataset, but they are not sufficient to
establish that the same limitations hold across all specialized or data-scarce
domains. Our conclusions should accordingly be interpreted as applying to the
evaluated sparse, high-variance classification settings rather than as a general
claim about synthetic data generation.

Second, the evaluated generative pipelines differ considerably in their
pretraining distributions, conditioning mechanisms, and sensitivity to
hyperparameters. Our objective was to evaluate these methods under standardized,
practitioner-oriented configurations rather than exhaustively optimize each
method independently. Consequently, stronger method-specific tuning,
alternative prompting strategies, or future generative models may produce
different results. This limits our ability to distinguish limitations inherent
to a particular generative approach from limitations of the specific
configuration evaluated here.

At the same time, visual artifacts, distributional drift, and failures to
preserve task-relevant semantics should not necessarily be regarded as
independent confounds in this setting. The ability to generate label-consistent,
domain-appropriate variation from a sparse seed dataset is itself part of the
augmentation problem studied here, and such failures constitute important
observed limitations of the evaluated pipelines. 

Third, our primary experiments use a frozen DINOv2 representation with an MLP classifier and emphasize macro-recall as the principal evaluation metric.
We partially test sensitivity to this choice through end-to-end ResNet50
experiments and synthetic-to-real mixing-ratio sweeps, which exhibit similar
qualitative trends. Nevertheless, these experiments do not cover the full space
of classifier architectures, representation-learning strategies, or evaluation
metrics. In particular, conclusions may differ under alternative backbones,
end-to-end representation learning, or metrics such as precision, F1-score,
balanced accuracy, or AUROC.

Our feature-space analyses introduce an additional representational dependency.
MMD and t-SNE are computed using DINOv2 embeddings, and therefore characterize
synthetic--real alignment as represented by that particular feature encoder.
Although these analyses provide evidence for distributional drift and clustering
behavior, distance in DINOv2 feature space is not equivalent to semantic or
task-relevant distance. Likewise, the observation that synthetic samples often
form simpler or more separable examples is consistent with our hardness
experiments, but does not uniquely establish the causal mechanism responsible
for downstream performance.

Finally, parts of our follow-up analysis rely on manually curated synthetic
exemplars selected for visual coherence and label consistency. This experiment
is intended to test whether removing obvious generation failures can recover
downstream utility, rather than to propose a scalable curation procedure.
Human selection introduces additional judgment and may bias the resulting
sample distribution. Developing automatic methods for assessing synthetic
sample validity, task relevance, and informativeness remains an important
direction for future work.

\subsection{Future Directions}

Our findings suggest several directions for improving robustness in sparse, high-variance domains.

From a data perspective, greater emphasis may be needed on structured collection protocols that reduce nuisance variance at acquisition time. This can be done at collection time by standardizing capture conditions, viewpoints, and annotations or via computational preprocessing strategies that suppress irrelevant variability—such as pose normalization, background removal, standardized relighting, or geometry-aware alignment.

Conversely, a generative perturber that systematically injects structured deployment artifacts—such as lighting variation, occlusion, motion blur, or viewpoint shifts—may better approximate the high-variance conditions encountered in real-world scenarios. The sample perturbation methods explored in this work already show promise in achieving this via a mix of text and image embedding prompting.

Explainability-driven analysis may provide a principled mechanism for improving synthetic data generation. Rather than evaluating synthetic images solely by visual plausibility, causal interpretability methods~\cite{chockler2024causal, navaratnarajah20253d} can identify the features and regions that actually drive a classifier’s predictions, revealing spurious shortcuts, overly canonical patterns, or semantically underrepresented cases. In this way, explainability can provide directional guidance for synthetic data generation.

Finally, it is important to note that improvements observed on held-out evaluation sets do not necessarily translate to increased robustness in real-world deployment. Thus, in safety-critical settings such as trauma detection, models must operate under uncontrolled environmental conditions that may differ substantially from both the original training data and synthetic augmentations. As a result, future work should emphasize evaluation protocols that better approximate deployment conditions and prioritize methods that improve robustness under real-world operational variability.

%% file: sections/8_acknowledgements.tex
\section*{Acknowledgments}

This research was partially supported by the DARPA Triage Challenge under
award HR00112420305. We gratefully acknowledge the collaborators who contributed to the data collection efforts supporting this work. Any opinions, findings, conclusions, or recommendations
expressed in this material are those of the authors and do not necessarily
reflect the views of DARPA or the U.S. Government.